\documentclass[letterpaper]{article}

\PassOptionsToPackage{table}{xcolor}
\usepackage{aaai2027}
\usepackage[hyphens]{url}
\usepackage{graphicx}
\usepackage{natbib}
\usepackage{caption}
\usepackage{booktabs}
\usepackage{amsmath}
\usepackage{amssymb}
\usepackage{amsthm}
\usepackage{multirow}
\title{OPTD: On-Policy Transition Distillation with Consistency-Guided Adaptive Compression for Few-Step Diffusion Language Models}

\author{
Xiaocheng Lu\textsuperscript{\rm 1,*},
Hualei Zhang\textsuperscript{\rm 1,*},
Shuhan Guo\textsuperscript{\rm 2,$\dagger$},
Jie Zhang\textsuperscript{\rm 1},
Xiaoyi Pang\textsuperscript{\rm 1},
Jian Liu\textsuperscript{\rm 1},\\
Haoxi Li\textsuperscript{\rm 1},
Bohai Gu\textsuperscript{\rm 1},
Haoxuan Che\textsuperscript{\rm 1},
Jingcai Guo\textsuperscript{\rm 3},
Song Guo\textsuperscript{\rm 1}
}
\affiliations{
\textsuperscript{\rm 1}HKUST \quad
\textsuperscript{\rm 2}NWPU \quad
\textsuperscript{\rm 3}PolyU
}
\newcommand{\authornotes}{%
\begingroup
\renewcommand{\thefootnote}{\fnsymbol{footnote}}%
\footnotetext[1]{Equal contribution.}%
\footnotetext[2]{Work completed during an internship at HKUST.}%
\endgroup}

\newcommand{\method}{\textsc{OPTD}}
\newcommand{\bestacc}[1]{\textbf{#1}}
\newcommand{\besttpf}[1]{\textbf{#1}}
\newcommand{\secondbest}[1]{\underline{#1}}
\newtheorem{lemma}{Lemma}
\nocopyright

\begin{document}

\maketitle
\authornotes

\begin{abstract}
Diffusion language models (dLLMs) can predict many tokens in parallel, but accurate generation still requires many iterative denoising steps. Few-step distillation accelerates decoding by compressing multiple teacher steps into a single student transition.
However, existing methods construct supervision on off-policy trajectories. At inference, the student's early parallel commitments alter the context of later predictions, so the states it actually visits drift away from the supervised ones---precisely when step compression is most aggressive.
On-policy distillation is a natural remedy for this mismatch, but it leaves open how far each transition should advance: matching only the teacher's next action limits compression, while indiscriminately merging future actions can violate intermediate dependencies.
To address this limitation, we propose OPTD, On-Policy Transition Distillation with consistency-guided adaptive compression. It samples partial states from the few-step student's own trajectories, uses a frozen, question-only teacher to identify outcome-aligned future candidates, and orders them by current-state confidence. The method then selects the longest prefix whose joint commitment preserves the teacher's rollout outcome. A set-bottleneck objective promotes every verified future candidate to the decoder's release threshold, while a frozen-teacher KL anchor regularizes all other active positions. Neither target construction nor training uses a gold response.
Across four mathematical reasoning and code-generation benchmarks, \method{}
consistently improves the quality--efficiency trade-off and attains the
strongest overall quality-constrained AUP among the evaluated few-step baselines.
\end{abstract}

\section{Introduction}

\begin{figure}[t]
\centering
\includegraphics[width=\columnwidth]{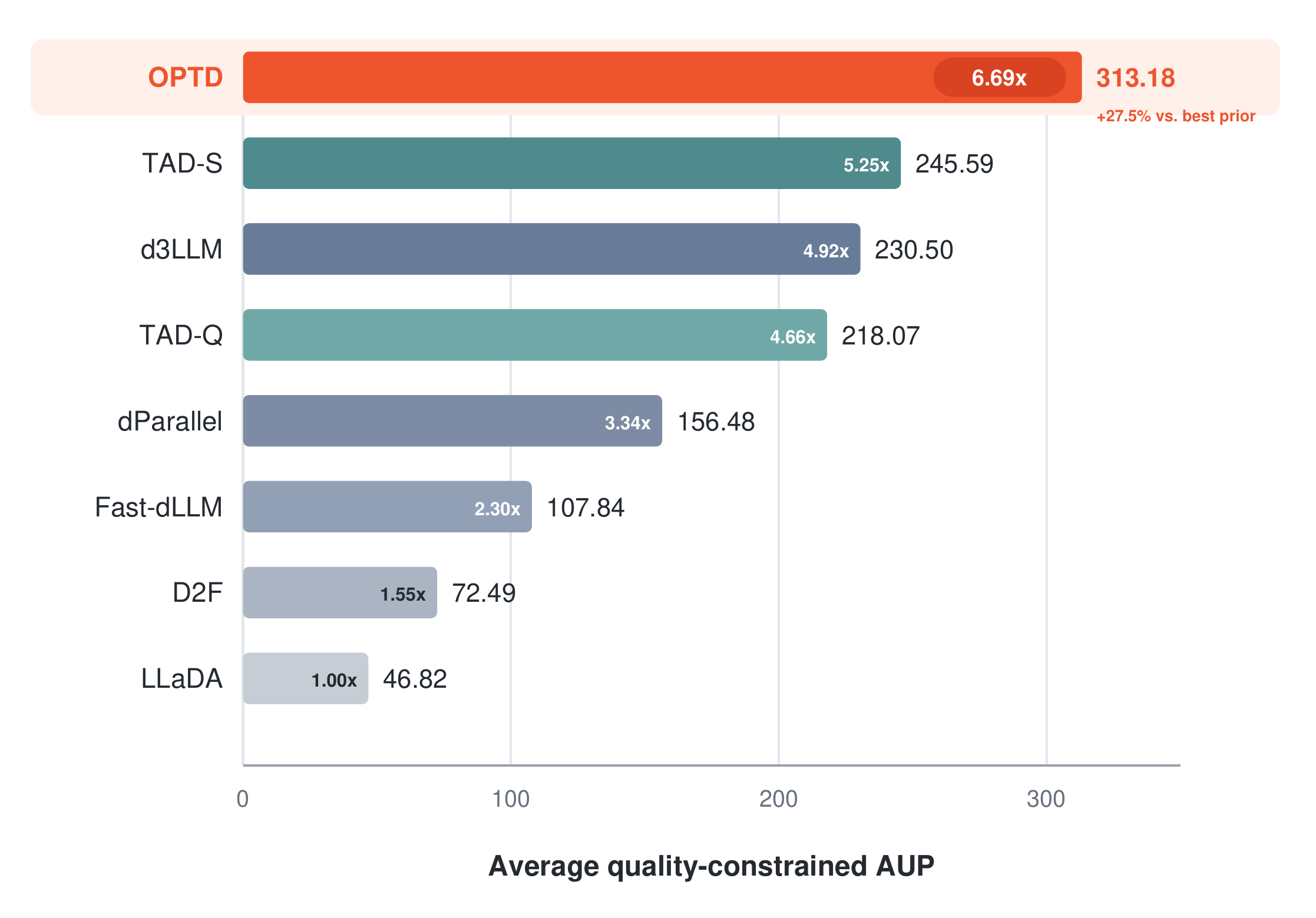}
\caption{\textbf{Quality--efficiency frontier.}
Average quality-constrained AUP across four benchmarks. \method{} achieves
313.18 AUP: 6.69$\times$ that of vanilla LLaDA and 27.5\% above the
highest-AUP prior method, TAD-S. The gain comes from higher parallelism with a
modest accuracy trade-off.}
\label{fig:main_aup}
\end{figure}

Diffusion language models (dLLMs) generate text by iteratively replacing
masked positions with predicted tokens~\citep{nie2025llada,sahoo2024mdlm}.
Because a denoiser predicts a distribution for every remaining mask, a decoder
can commit multiple confident tokens in parallel. Accurate generation
nevertheless requires many neural function evaluations (NFEs), since later
predictions often depend on context revealed by earlier denoising steps.
Few-step distillation reduces this cost by training one student transition to
approximate several teacher transitions.
Figure~\ref{fig:main_aup} previews the resulting four-benchmark
quality--efficiency trade-off.

Most existing few-step methods construct compression targets off-policy, using
fixed teacher trajectories or prescribed corruption schedules rather than the
accelerated student's own behavior~\citep{t3d2026, d3llm2026, cdlm2026,
cd4lm2026, speed2026, tad2026}. At inference, the student determines both the
tokens committed and the next partial state. This mismatch is particularly
consequential for dLLMs because one parallel action changes the context of all
remaining masks and can alter many later release decisions. In our audit, only
33.7\% of student-visited states occur on the corresponding teacher trajectory;
coverage falls from 55.0\% in the first eight decoding steps to 6.5\% after
step 40.
Moreover, only 50.3\% of merges verified on progress-matched teacher states
remain valid after transfer to student states.

On-policy distillation aligns supervision with student-visited states, but it
does not determine how far a transition should advance. Matching only the
teacher's immediate action provides no additional compression, whereas a fixed
merge horizon imposes the same depth at every state and, in our horizon
diagnostic, fails to convert deeper horizons into additional parallelism.
Token-wise verification is also insufficient: a future token may match its value
in the teacher outcome in isolation yet become invalid when committed jointly
with other future tokens. Indeed, we later find that unions of individually
valid tokens can still fail the joint check (10 of 416 in our verification-unit
ablation). These two failure modes motivate a state-dependent compression rule
evaluated directly on student-visited states.

We propose \method{}, On-Policy Transition Distillation with
Consistency-Guided Adaptive Compression. \method{} samples partial states from
the evolving student's inference policy and queries a frozen, question-only
teacher (it sees only the question and the current partial state) for
outcome-aligned future candidates. It orders these candidates by current-state
confidence and selects the longest prefix whose joint commitment preserves the
teacher's rollout outcome. A set-bottleneck certainty loss promotes every
verified future token to the decoder's release threshold, while a frozen-teacher
KL anchor regularizes all other active positions. The teacher and verifier are
used only to construct training targets; inference runs the unchanged student
and decoder. Empirically, this procedure improves decoding parallelism while
maintaining competitive accuracy.
No model or target-construction component observes a gold response or reference
answer.

In summary, our main contributions are threefold:
\begin{itemize}
    \item We identify and quantify the state--action mismatch in off-policy
    few-step dLLM distillation, showing how teacher-collected supervision
    diverges from the states and release decisions encountered by the evolving
    student.
    \item We introduce consistency-guided adaptive compression, which selects
    the longest outcome-preserving candidate prefix and optimizes it with
    threshold-aligned verified-set certainty forcing and a frozen-teacher
    anchor.
    \item Across mathematical reasoning and code-generation benchmarks,
    \method{} improves accuracy--parallelism trade-offs over few-step
    baselines; controlled analyses isolate the effects of on-policy states and
    joint consistency verification.
\end{itemize}

\section{Related Work}

\paragraph{Diffusion language models.}
Discrete diffusion extends denoising to categorical state
spaces~\citep{austin2021d3pm,hoogeboom2021argmax}, and recent masked dLLMs
scale this formulation to language modeling~\citep{sahoo2024mdlm,
arriola2025blockdiffusion,nie2025llada,ye2025dream,cheng2025sdar}. At a
partial state $x$, a denoiser predicts a distribution for every masked
position. A dynamic decoder releases positions whose confidence exceeds a
threshold, with a fallback release when no position is ready. Therefore, model
quality and confidence calibration jointly determine the number of tokens
released per forward. Fast-dLLM, D2F, and dParallel accelerate this decoding
process through prefix caching and parallel release mechanisms~\citep{
wu2025fastdllm,wang2026d2f,chen2025dparallel}; these system-level optimizations
are complementary to the learned transition policy.

\paragraph{Few-step and trajectory distillation.}
Progressive distillation and consistency learning compress multi-step diffusion
samplers in continuous domains~\citep{salimans2022progressive,
song2023consistency}. Recent dLLM methods use pseudo-trajectories, temporal
targets, sharpened distributions, learned release order, or consistency
objectives to reduce denoising cost~\citep{d3llm2026,cdlm2026,cd4lm2026,
speed2026,tad2026,t3d2026}. These methods establish strong distilled
checkpoints, but their supervision is generally produced before or
independently of the student's inference-time trajectory. \method{} is
compatible with these initialization recipes and instead contributes
inference-aligned state collection and adaptive transition compression. Our
notion of consistency also differs: it is outcome consistency of the completed
block under a jointly committed merge, rather than self-consistency along a
sampler trajectory.

\begin{figure*}[t]
\centering
\includegraphics[width=\textwidth,pagebox=cropbox]{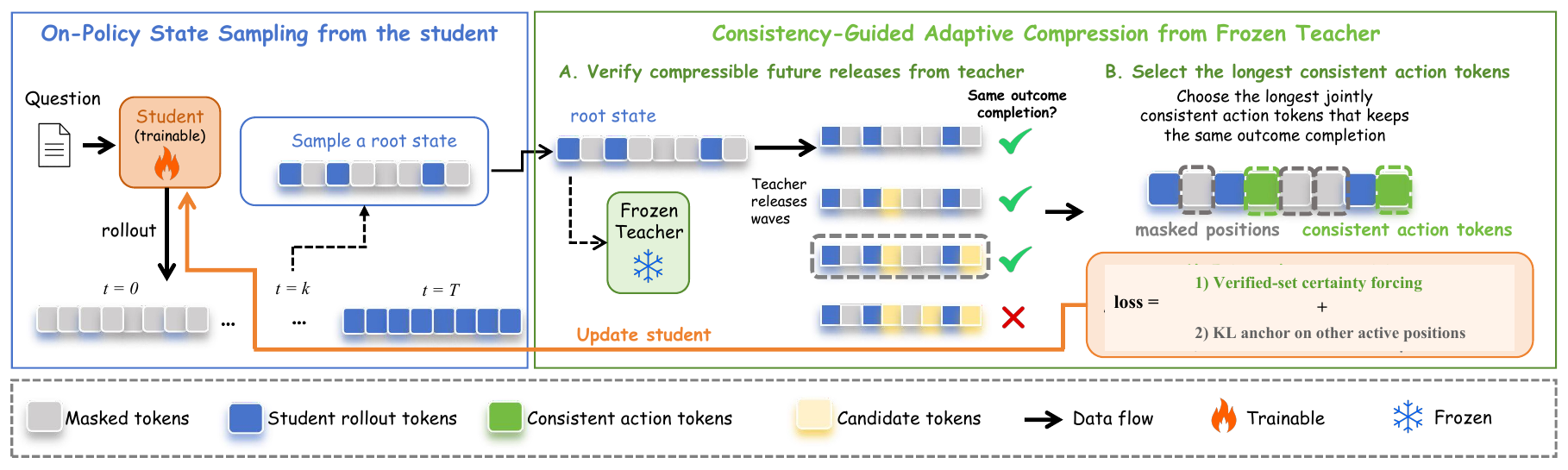}
\caption{\textbf{Overview of \method{}.}
The current student supplies on-policy partial states. A frozen, question-only
teacher rolls forward from each state and proposes future releases. \method{}
retains candidate actions whose joint commitment preserves the teacher's
rollout outcome and selects the longest outcome-preserving action.
Threshold-aligned set-bottleneck certainty forcing promotes its verified
future tokens, and a frozen-teacher KL anchor regularizes all other active
positions.}
\label{fig:optd_framework}
\end{figure*}

\paragraph{On-policy distillation.}
The state-distribution mismatch between teacher and student is the core motivation of on-policy distillation: GKD lets an autoregressive student generate its own trajectories and queries a teacher on them to obtain dense next-token supervision~\citep{agarwal2023gkd}. dLLMs exhibit the same mismatch, motivating a line of on-policy distillation for this setting: OPDLM performs data-efficient knowledge transfer from an autoregressive teacher to a diffusion student, yet supervises only final sequences and leaves intermediate denoising states unconstrained~\citep{su2026opdlm}; d-OPSD refines on-policy self-distillation to the step level by supervising arbitrary-order release transitions, but relies on privileged self-future conditioning that is unavailable at inference~\citep{luo2026dopsd}; and TOPD conducts trace-based OPD along the student's own denoising trajectory~\citep{ren2026topd}. All three align teacher supervision with the student's visited states, but pursue capability or alignment gains rather than few-step acceleration. \method{} instead learns accelerated merges directly on student states.

\section{On-Policy Transition Distillation}

\subsection{Preliminaries}

A masked dLLM starts from a fully masked response. Given question $q$, one
denoiser forward at partial state $x$ predicts
$p_\theta(\cdot\mid q,x,i)$ at every active mask $i\in M(x)$. Let $G(x)$ be
the currently exposed block group and $M_G(x)\subseteq M(x)$ its masked
positions. Define $c_\theta(i\mid q,x)=\max_y p_\theta(y\mid q,x,i)$. A
threshold decoder selects
\begin{equation}
A_\theta^\tau(q,x)
=\{i\in M_G(x):c_\theta(i\mid q,x)\geq\tau\}
\end{equation}
and jointly commits their argmax tokens, with a most-confident fallback for an
empty set. We count each denoiser forward as one NFE.

Few-step distillation replaces several teacher transitions with one student
transition. Existing methods train on a fixed state distribution
$d_{\mathrm{off}}$, whereas inference follows the student's own distribution $d_\theta$:
\begin{equation}
\mathcal{L}_{\mathrm{off}}=\mathbb{E}_{x\sim d_{\mathrm{off}}}\ell(x;\theta),
\qquad
\mathcal{L}_{\mathrm{on}}=\mathbb{E}_{x\sim d_\theta}\ell(x;\theta).
\end{equation}
Since a joint release alters the context of all remaining masks
simultaneously, targets constructed on $d_{\mathrm{off}}$ often become invalid
under $d_\theta$. To address this distributional mismatch, \method{} optimizes
under the student-induced state distribution by constructing supervision
directly on partial states sampled as $x\sim d_\theta$.

\subsection{The \method{} Algorithm}

Although on-policy training resolves the state-distribution mismatch, it does
not determine the appropriate compression extent: how many teacher transitions
can be consistently merged into a single student transition at a given state?
\method{} resolves this state-dependent compression problem through adaptive
transition construction. As illustrated in Figure~\ref{fig:optd_framework},
the current student first supplies an on-policy state using its inference
decoder. We call this sampled student-visited state the \emph{root state}. A
frozen teacher then generates a conservative continuation from that state.
From this continuation, \method{} selects the longest
confidence-ordered future action whose joint commitment preserves the teacher's rollout outcome.

\paragraph{On-policy state collection.}
Given a training question $q$, we run the current student from a fully masked
response to a sampled rollout depth using the same question-only dynamic
decoder as at inference, retaining the resulting partial state $x$ and the
block group exposed to its next transition. We then query a frozen teacher
policy $p_T$ with $(q,x)$ to obtain its current
distribution and a conservative future continuation.
% It does not observe a gold response, reference answer, or future student
% tokens. Freezing $p_T$ provides a stable construction policy and anchor
% throughout refinement. The extra rollout and counterfactual completion are
% training-only and expose no privileged information to either model.

\paragraph{Consistency-guided adaptive compression.}
To determine how far the teacher trajectory can be safely compressed at each state, we perform consistency-guided adaptive compression: we construct
an ordered family of outcome-aligned future candidates and select a
state-dependent prefix length by testing their effect on the frozen teacher's rollout outcome.

Specifically, starting from a student-visited state $x^0=x$, the frozen teacher
uses the confidence threshold $\tau$ to complete the currently exposed block
group conservatively:
\begin{equation}
x^0 \xrightarrow{A_1} x^1 \xrightarrow{A_2} \cdots
\xrightarrow{A_J} x^J=x^T ,
\end{equation}
where $A_h$ is the release set at teacher wave $h$ and $x^T$ is the rollout
outcome. The first wave $A_1$ is immediately available at $x$; later waves
generally condition on the intervening releases. The exposed block group bounds
the spatial scope of target construction; within that group, the method imposes
no fixed per-state compression rate. The teacher-wave horizon $H$ is bounded
by both the exposed block group and an implementation cap of $H{=}16$ waves.

Let $\hat y_i=\arg\max_y p_T(y\mid q,x,i)$ and
$c_i=p_T(\hat y_i\mid q,x,i)$. A future position is eligible only if
its current-state teacher prediction already matches its value in the teacher outcome:
\begin{equation}
\mathcal{C}(x)=
\left\{i\in M_G(x)\setminus A_1:\;\hat y_i=x^T_i\right\},
\end{equation}
where $M_G(x)$ restricts candidates to the currently exposed block group. We
order candidates by decreasing $c_i$, using position as a deterministic
tie-breaker, and let $C_m$ be the first $m$ candidates. Teacher release waves
are used to construct the rollout outcome and to measure the wave span of
accepted candidates (for reporting), but do not partition or order the
candidate set. The identity filter removes candidates whose current
predictions still change along the teacher trajectory, while confidence
ordering prioritizes candidates already closest to the release boundary.

However, satisfying this token-wise identity filter is insufficient for parallel decoding.
Current agreement is a token-wise condition: it identifies
candidate tokens whose identities appear stable, but cannot establish that
several candidates are jointly independent of the teacher's intermediate
states. Committing them together changes the context seen by every remaining
mask.
To resolve this, we evaluate each proposed action as a set and reject a merge
whenever it changes the teacher's rollout outcome $x^T$. Let
$\Phi_T(x;B)=\operatorname{Outcome}_T(\operatorname{Commit}(x,B))$ denote the
frozen teacher's rollout outcome after jointly committing $B$ at $x$.
Beginning with the longest prefix, we select
\begin{equation}
m^\star=\max\{m\in\{0,\ldots,|\mathcal{C}(x)|\}:
\Phi_T(x;A_1\cup C_m)=x^T\}.
\label{eq:outcome_equivalence}
\end{equation}
Equality is token-wise over every exposed block, so candidates may cross a
block boundary without changing the criterion. Here $C_0=\varnothing$, so the
ordinary first teacher release is always the fallback. The verified future set
is $V(x)=C_{m^\star}$ and the compressed action is
$S(x)=A_1\cup V(x)$.

\subsection{Outcome-Consistency Guarantee}

\begin{lemma}[Fallback consistency and verification cost]
\label{lem:consistent_prefix}
Assume that the frozen teacher and its tie-breaking rules are deterministic.
Let $n=|\mathcal{C}(x)|$. Then $C_0=\varnothing$ is consistent, so $m^\star$
in \eqref{eq:outcome_equivalence} is well-defined and the returned action is
jointly outcome-preserving:
\[
\Phi_T(x;S(x))=x^T.
\]
A descending exhaustive scan finds $m^\star$ using at most $n$ nontrivial
counterfactual outcome checks, or $\lceil n/b\rceil$ verifier batches when
$b$ prefixes are checked in parallel; $C_0$ needs no check because it is the
ordinary first teacher transition. This does not assume that outcome consistency
is monotone in $m$.
\end{lemma}

By definition, $C_{m^\star}$ is the longest, and hence largest-cardinality,
outcome-preserving member of the prescribed prefix family. This observation is
prefix-relative: it does not claim optimality over arbitrary subsets of
$\mathcal{C}(x)$, whose exhaustive search has $2^n$ candidates. The ordering
prioritizes current-state teacher confidence while making exact verification
tractable. The proof of Lemma~\ref{lem:consistent_prefix} and a
formal statement of its scope are given in the technical appendix
(supplementary material). We construct this single verified prefix action at
the student-visited root state;
teacher-generated intermediate states are used for verification but are not
added to the training distribution.

\paragraph{Scope of the guarantee.}
Lemma~\ref{lem:consistent_prefix} concerns the construction of $S(x)$ under
the frozen teacher and verifier. At inference, neither component is present:
the student releases $A_\theta^\tau(q,x)$ according to its learned
probabilities, and this set need not equal $S(x)$ or be outcome-preserving
under the teacher. Accordingly, the lemma establishes outcome-preserving
training targets under the frozen teacher, not an inference-time consistency
guarantee; inference quality remains an empirical property
of the learned policy and its confidence calibration.

\subsection{Verified-Set Transition Objective}

Target construction verifies token identities and their joint outcome effect.
The ordinary first wave $A_1$ is already above the frozen teacher's release
threshold; the additional verified set $V(x)$ is what must be promoted to make
the merged transition available in one student step.  Let
$h_i(\theta)=[\log\tau-\log p_\theta(\hat y_i\mid q,x,i)]_+$ and use
\begin{equation}
\mathcal{L}_{\mathrm{set}}
=
\begin{cases}
\max_{i\in V(x)}h_i(\theta), & V(x)\neq\varnothing,\\
0, & V(x)=\varnothing.
\end{cases}
\end{equation}
It is zero only when every verified candidate reaches the release boundary
and otherwise targets the least-ready member. All other active positions,
including the already releasable first wave, retain the frozen teacher's
current-state distribution:
\begin{equation}
\mathcal{L}_{\mathrm{anchor}}
=\frac{1}{|M_G(x)\setminus V(x)|}
\sum_{i\in M_G(x)\setminus V(x)}
D_{\mathrm{KL}}\!\left(p_T^i\,\|\,p_\theta^i\right),
\end{equation}
with zero value for an empty complement. Overall,
$\mathcal{L}_{\mathrm{OPTD}}=\mathcal{L}_{\mathrm{set}}
+\lambda_{\mathrm{anchor}}\mathcal{L}_{\mathrm{anchor}}$; the primary
checkpoint uses $\lambda_{\mathrm{anchor}}=1$. No auxiliary head, gold target, or inference-time
verifier is required.

\paragraph{Mean-hinge variant.}
For the loss-design ablation, we also replace the set bottleneck by
$|V|^{-1}\sum_{i\in V}h_i$, with zero value for an empty set, while retaining
the identical teacher anchor. Mean-hinge improves average
readiness; set-bottleneck targets the least-ready token in each joint release.

\section{Experiments}

\begin{table*}[t]
\centering
\caption{Full-benchmark complete-stack results using each method's native
decoder; MBPP uses all 500 test problems. AUP anchors the conservative endpoint
at algorithmic TPF $=1$ and may use a quality-constrained threshold-sweep
endpoint from the same decoder family (technical appendix). The displayed
\method{} operating point uses confidence threshold $\tau=0.8$. $^\dagger$:
conservative endpoint; TPF is 1.00 by definition. Best results are bold,
second-best are underlined, and our method is shaded.}
\label{tab:main_algorithms}
\small
\setlength{\tabcolsep}{1.8pt}
\renewcommand{\arraystretch}{1.08}
\resizebox{\textwidth}{!}{%
\begin{tabular}{l*{5}{ccc}}
\toprule
\multirow{2}{*}{Method}
& \multicolumn{3}{c}{GSM8K}
& \multicolumn{3}{c}{MATH-500}
& \multicolumn{3}{c}{HumanEval}
& \multicolumn{3}{c}{MBPP}
& \multicolumn{3}{c}{Average} \\
\cmidrule(lr){2-4}\cmidrule(lr){5-7}\cmidrule(lr){8-10}\cmidrule(lr){11-13}\cmidrule(l){14-16}
& Acc. & TPF & AUP & Acc. & TPF & AUP & Acc. & TPF & AUP & Acc. & TPF & AUP & Acc. & TPF & AUP \\
\midrule
LLaDA (vanilla)$^\dagger$~\citep{nie2025llada}
& 76.50 & 1.00 & 76.50
& 34.20 & 1.00 & 34.20
& 36.59 & 1.00 & 36.59
& 40.00 & 1.00 & 40.00
& 46.82 & 1.00 & 46.82 \\
Fast-dLLM~\citep{wu2025fastdllm}
& 79.53 & 3.25 & 256.47
& 35.60 & 2.14 & 69.18
& 37.80 & 2.50 & 56.79
& 38.60 & 2.25 & 48.93
& 47.88 & 2.54 & 107.84 \\
D2F~\citep{wang2026d2f}
& 68.54 & 2.97 & 161.92
& 20.40 & 2.75 & 26.25
& 32.32 & 2.73 & 50.41
& 38.80 & 2.34 & 51.37
& 40.01 & 2.70 & 72.49 \\
dParallel~\citep{chen2025dparallel}
& 75.74 & 5.85 & 392.77
& 30.60 & 4.46 & 79.57
& 37.20 & \secondbest{7.50} & 97.73
& 39.80 & 2.36 & 55.84
& 45.84 & 5.04 & 156.48 \\
d3LLM~\citep{d3llm2026}
& 76.35 & \secondbest{8.62} & \secondbest{616.04}
& 30.00 & \secondbest{6.61} & 116.77
& 35.98 & 7.07 & 106.99
& 39.20 & \besttpf{5.46} & 82.20
& 45.38 & \secondbest{6.94} & 230.50 \\
TAD-Q~\citep{tad2026}
& \secondbest{79.76} & 6.29 & 481.72
& \bestacc{46.80} & 4.41 & \secondbest{197.39}
& \secondbest{40.24} & 6.06 & 108.54
& \bestacc{40.80} & 3.75 & \textbf{84.61}
& \bestacc{51.90} & 5.13 & 218.07 \\
TAD-S~\citep{tad2026}
& \bestacc{80.21} & 8.34 & 576.34
& \secondbest{40.80} & 5.10 & \textbf{216.26}
& \bestacc{40.85} & 5.94 & \secondbest{115.53}
& \secondbest{40.60} & 3.42 & 74.23
& \secondbest{50.62} & 5.70 & \secondbest{245.59} \\
\rowcolor{gray!10}
\textbf{\method{}}
& 78.85 & \besttpf{10.85} & \textbf{848.31}
& 34.00 & \besttpf{7.41} & 185.19
& 37.20 & \besttpf{9.24} & \textbf{136.67}
& 38.80 & \secondbest{3.98} & \secondbest{82.56}
& 47.21 & \besttpf{7.87} & \textbf{313.18} \\
\bottomrule
\end{tabular}
}
\end{table*}

\subsection{Setup}

We initialize student and frozen teacher from the released TAD-Speed checkpoint
(denoted TAD-S throughout), based on
LLaDA-8B-Instruct~\citep{nie2025llada,tad2026}. The primary checkpoint uses
512 two-rank updates and rank-128, alpha-128 LoRA on all linear projections
with learning rate $10^{-6}$~\citep{hu2021lora}. It
collects student states and constructs maximal verified future sets with a
frozen teacher. Training uses the set-bottleneck certainty loss and residual
teacher KL defined above. Rollout and deployment both use confidence
Multi-Block decoding with response length 256, block length 32, at most $K=3$
active blocks, confidence threshold $\tau=0.8$, and block-add and decoded-token
thresholds $0.5$. This terminal checkpoint supplies the
\method{} entries in Tables~\ref{tab:main_algorithms}
and~\ref{tab:multiblock_compat}. The single-block attribution studies and
the separate $H=16$ compression-space audit use different configurations,
detailed in the technical appendix.

We evaluate on GSM8K (5-shot), MATH-500 (4-shot), MBPP (3-shot), and HumanEval
(0-shot)~\citep{cobbe2021gsm8k,hendrycks2021math,
chen2021humaneval,austin2021programsynthesis}. We use BF16, greedy decoding,
batch size 1, and a 256-token budget. We report exact-match accuracy or pass@1, NFE, and
tokens per forward (TPF). Following d3LLM, the unaccelerated endpoint is
anchored at TPF $=1.00$; accelerated endpoints use the number of emitted
non-special tokens divided by the actual NFE. All MBPP entries use the same 500-problem
denominator; generated code is extracted, sanitized, and executed with an
isolated pass@1 scorer. As in d3LLM, AUP uses $\alpha=3$, an unweighted
TPF-$1$ accuracy anchor, and trapezoidal integration of quality-weighted
accuracy beyond that anchor. We retain only endpoints within five accuracy
points of the conservative endpoint and average AUP across tasks; the
technical appendix gives the exact formula, reference accuracies, and endpoint
policy.

\subsection{Main Results}

Table~\ref{tab:main_algorithms} compares method-native inference stacks under
a common evaluation protocol (prompts, scorers, and budgets) across four
benchmarks. It is therefore a complete-stack comparison, not an isolation of
transition distillation; the controlled ablations in the next subsection
provide the latter attribution.

\paragraph{Aligned cross-task comparison.}
\method{} reaches 47.21\% average accuracy and 7.87 average TPF. As an
end-to-end stack, it improves over d3LLM by 1.83 accuracy points and 13.4\% in
TPF. Relative to TAD-S, \method{} increases TPF by 38.1\% in exchange for a
3.41-point accuracy drop. These complete-stack deltas include different native
decoders and must not be attributed solely to transition distillation.
Table~\ref{tab:multiblock_compat} therefore provides a decoder-matched
transition-policy attribution across all four benchmarks. \method{}'s average
accuracy is 0.39 points above vanilla LLaDA (46.82\%), while its average TPF is
$7.87\times$ vanilla LLaDA's anchored TPF. Thus, \method{} is a
high-parallelism quality--efficiency operating point rather than a uniformly
accuracy-preserving method; it achieves the highest TPF on GSM8K, MATH-500,
and HumanEval.

\paragraph{Quality-constrained AUP.}
The Acc/TPF columns in Table~\ref{tab:main_algorithms} show each method's
primary operating point, whereas AUP summarizes its quality-constrained
full-dataset curve. The technical appendix specifies when the curve uses a
quality-constrained threshold-sweep endpoint from the same decoder family.
\method{} obtains the highest average AUP
(313.18), compared with 245.59 for TAD-S, 230.50 for d3LLM, and 218.07 for
TAD-Q. It leads on GSM8K and HumanEval; TAD-S leads on MATH-500, and TAD-Q
leads on MBPP.

\paragraph{Complete inference stacks.}
On full GSM8K with each method's released inference strategy, TAD-S reaches
81.35\%/8.512 Acc/TPF and d3LLM 72.71\%/9.212, close to their reported
operating points. Policy-matched K3 reaches 78.85\%/10.805: it exceeds d3LLM
on both metrics and offers a higher-parallelism point than TAD-S. Decoder,
cache, sharding, and hardware details are in the technical appendix.

\subsection{Ablation Experiments}

We first isolate the learned transition policy from the inference stack, then
ablate its state source, teacher update, set reduction, and verification unit.
Each comparison changes only the named factor and otherwise matches the
evaluator, training budget, and decoding configuration.

\paragraph{Transition-policy attribution.}
We evaluate d3LLM and set-bottleneck \method{} on the same full-GSM8K examples
with one evaluator, a 256-token generation budget, and a single-block
prefix-cache decoder. Under this matched setting, \method{} improves accuracy
from 75.13\% to 79.91\% and TPF from 6.842 to 6.995. It corrects 157 examples
that d3LLM fails and regresses on 94 examples (exact McNemar
$p=8.41\times10^{-5}$). Average NFE is essentially unchanged (34.74 versus
34.87), while \method{} outputs are 2.62\% longer. The TPF gain therefore
reflects more decoded tokens per forward rather than fewer forwards at this
operating point; we do not claim an NFE reduction.

\paragraph{State source.}
We keep the TAD-S initialization, frozen teacher, verifier, loss, update
budget, and decoder fixed. The teacher-state control
applies the identical objective to frozen-teacher states; \method{} instead
trains on student-visited states. Both use rank-128 LoRA, learning rate
$10^{-6}$, and a confidence Multi-Block decoder with $K=3$ and $\tau=0.8$.

\begin{figure}[t]
\centering
\includegraphics[width=\columnwidth]{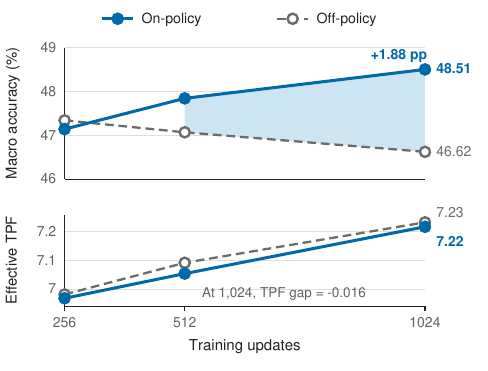}
\caption{Single-seed matched state-source learning curves. Accuracy is the macro average
over full GSM8K, MATH-500, MBPP-500, and HumanEval; TPF uses the same decoder.
The on-policy accuracy gap grows with training while efficiency remains
matched.}
\label{fig:state_source_curve}
\end{figure}

Figure~\ref{fig:state_source_curve} provides the state-source attribution. On-policy
macro accuracy improves monotonically from 47.14\% to 47.85\% to 48.51\%
across 256/512/1,024 updates. Its gap over off-policy changes from $-0.20$ to
$+0.78$ to $+1.88$ points, whereas final macro TPF is nearly identical
(7.217 versus 7.233). At 1,024 updates, on-policy gains 0.45/3.00/4.88 points
on GSM8K/MATH-500/HumanEval and loses 0.80 on MBPP-500. The observed aggregate
gap becomes positive only after sufficient optimization; we do not claim
dominance at every checkpoint or on every task. This single-seed curve provides
matched learning-curve evidence rather than a variance estimate.

\paragraph{Teacher update.}
Refreshing the teacher to match the current student before every update yields
79.98\%/6.285 Acc/TPF, versus 81.05\%/6.171 for the frozen teacher at matched
checkpoints. Self-teaching collapses the residual anchor KL to
$4.66\times10^{-9}$, leaving only future-release sharpening; the frozen
teacher maintains an anchor KL of 0.00697 and prevents recursive certainty
drift (the student teaching itself ever-higher confidence) on student-visited
states during training.

\paragraph{Set reduction.}
\begin{table}[t]
\centering
\caption{Matched loss ablation at 1,024 updates and $\tau=0.8$. Benchmark
columns report full-set accuracy; Avg. and TPF are four-benchmark macro means.}
\label{tab:set_bottleneck}
\footnotesize
\setlength{\tabcolsep}{1.2pt}
\renewcommand{\arraystretch}{0.9}
\resizebox{\columnwidth}{!}{%
\begin{tabular}{@{}lrrrrrr@{}}
\toprule
Loss & GSM & MATH & MBPP & HE & Avg. & TPF \\
\midrule
Mean & 76.57 & 35.00 & 34.80 & 37.80 & 46.04 & \textbf{7.44} \\
Proj.\ KL & 76.80 & 33.60 & 37.20 & 37.20 & 46.20 & 7.19 \\
Set bottleneck (ours) & \textbf{77.86} & \textbf{36.20} & \textbf{37.40}
  & \textbf{39.02} & \textbf{47.62} & 7.20 \\
\bottomrule
\end{tabular}
}
\end{table}

Table~\ref{tab:set_bottleneck} shows that mean hinge gains 0.24 TPF but loses
1.58 accuracy points; projection KL
(forward KL to a teacher distribution whose verified targets are lifted to the
release boundary; technical appendix) loses 1.42 points at similar speed. Since
the least-ready member controls joint release, we adopt the max hinge as our
primary, accuracy-oriented objective. Its early disadvantage at 256 updates
reverses by 512 (technical appendix), so we do not claim dominance at every
operating point. In the earlier single-block recipe, set bottleneck at
evaluation threshold 0.85 gains 0.83 accuracy points over mean hinge at a
1.47\% TPF cost; at the nearest-speed operating points the paired difference is
not significant ($p=0.930$; technical appendix).

\paragraph{Verification unit.}
Independent verification admits more tokens per state (2.875 versus 1.327),
but 10 of 416 unions of individually valid tokens fail the joint check. Joint testing gains
0.67/0.89 accuracy points at updates 256/512, whereas independent testing gains
0.16/0.26 TPF. Thus joint testing is the appropriate unit of verification;
independent testing trades quality for speed.

\subsection{Analysis Experiments}

Having isolated the design choices above, we next analyze why on-policy
collection is needed, how much state-dependent compression is available,
whether its gain survives a decoder-matched comparison, and how much of the
compression benefit is attributable to adaptive scheduling alone. These
studies diagnose the mechanism and its operating regime; the final one
isolates scheduling adaptivity outside the full \method{} stack.

\paragraph{On-policy distribution shift.}
We show that the states visited by the accelerated student drift away from
frozen-teacher trajectories, so purely offline supervision covers them poorly.
We sample one state uniformly from each complete rollout and compare it against
every state of the corresponding frozen-teacher
trajectory. Only 33.7\% are covered (95\% Wilson CI $[30.9,36.6]$). Coverage
decreases from 55.0\% during student steps 0 through 7 to 6.5\% after step 40
($r=-0.317$, permutation $p<10^{-4}$). Meanwhile, normalized
generated-sequence Hamming distance to the nearest teacher state increases from
0.0119 to 0.4510 ($r=0.561$, $p<10^{-4}$). This confirms that teacher
trajectories provide poor support for student-visited states.

\begin{figure*}[t]
\centering
\includegraphics[width=\textwidth]{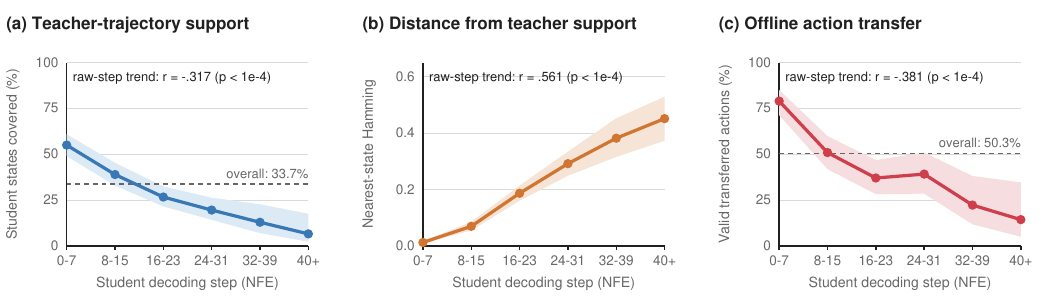}
\caption{Student states leave teacher support (a,b), while transferred teacher
actions become unreliable (c) over matched rollouts; shading gives 95\%
intervals.}
\label{fig:onpolicy_motivation}
\end{figure*}

For action comparison, we select a teacher state in the same current block by
remaining-mask count. Among sampled states, 61.2\% have an exactly
count-matched teacher state, and 75.6\% differ by at most one mask. We then run
the frozen teacher and the same outcome-consistency verifier on both states.
Figure~\ref{fig:onpolicy_motivation} shows that progress matching does not make
their actions interchangeable.

Among teacher states for which the verifier selects a nonempty future set,
only 50.3\% remain outcome-preserving after transfer to the paired
student state. Validity falls from 78.9\% during steps 0 through 7 to 14.3\%
after step 40 ($r=-0.381$, permutation $p<10^{-4}$). The frozen
teacher is also less certain on student states: mean confidence decreases by
0.0170 (95\% bootstrap CI $[-0.0262,-0.0080]$), and the fraction above the 0.9
release threshold decreases by 4.97 points (CI $[-7.02,-3.01]$). The shift
affects release calibration and action validity rather than introducing
harmless progress noise.

\paragraph{Verified compression space.}
On student-visited states at threshold 0.9, the mean of
$|A_1|$, $|V(x)|$, and $|S(x)|$ is 9.43, 4.52, and 13.94, and the mean
verified horizon is 2.41 waves; $|\mathcal{C}(x)|=0$ in 31.4\% of states.
The full diagnostic in the technical appendix shows that a requested
8$\times$ span is fully available in only 10.1\% of states, and the realized
span adaptively caps at 3.16$\times$; at a requested 16$\times$, utilization
falls to 21.2\%. A rank-256 LoRA yields essentially the same cap
(3.17$\times$). Teacher-wave span should not be confused with TPF: under the
matched decoder, \method{} reaches 6.995 TPF against the practical single-block
ceiling of 8 (32-token blocks decoded in at least 4 forwards).

The single-block frozen-teacher run accepts 42.5\% of states and promotes 1.33
future tokens per state.

\paragraph{Decoder-matched policy attribution.}
We apply the same confidence Multi-Block decoder to released TAD-S and
\method{} on all four full benchmarks.

\begin{table}[t]
\centering
\caption{Decoder-matched full-benchmark results under confidence Multi-Block
decoding (no explicit active-block cap, $\tau=0.8$); TAD-S has no \method{}
adapter.}
\label{tab:multiblock_compat}
\small
\setlength{\tabcolsep}{3.0pt}
\renewcommand{\arraystretch}{0.9}
\begin{tabular}{lrrrr}
\toprule
& \multicolumn{2}{c}{TAD-S} & \multicolumn{2}{c}{\method{}} \\
\cmidrule(lr){2-3}\cmidrule(lr){4-5}
Benchmark & Acc. & TPF & Acc. & TPF \\
\midrule
GSM8K     & 77.79 & 9.74 & \textbf{77.86} & \textbf{10.04} \\
MATH-500  & 34.40 & 6.66 & \textbf{37.20} & \textbf{6.96} \\
HumanEval & 37.80 & 7.18 & \textbf{38.41} & \textbf{7.66} \\
MBPP-500  & \textbf{39.20} & 3.87 & 38.40 & \textbf{3.99} \\
\midrule
Average   & 47.30 & 6.86 & \textbf{47.97} & \textbf{7.16} \\
\bottomrule
\end{tabular}
\end{table}

Under the identical decoder and threshold, \method{} raises macro accuracy
from 47.30 to 47.97 ($+0.67$ points) and average TPF from 6.86 to 7.16
($+4.4\%$), with per-task TPF gains of 3.0--6.7\%. The shared decoder supplies base cross-block parallelism, but the
learned policy further improves release efficiency; accuracy does not improve
universally.

\paragraph{Transfer from base LLaDA.}
We also test whether useful compression can be learned without the TAD-S
initializer. Starting from LLaDA-8B-Instruct with no adapter, we train a
rank-128 adapter for 320 updates. A frozen base model, decoded with Fast-dLLM
at confidence 0.9, supplies teacher trajectories. This stress test changes
multiple training and decoding factors relative to the primary model, so it
is evidence of transfer rather than a controlled initializer ablation.

\begin{table}[t]
\centering
\caption{Transfer from base LLaDA. Vanilla is the TPF-$1$ conservative
decode of the trained checkpoint; ``fast valid'' reports Acc./TPF at each
task's fastest endpoint within the five-point quality constraint.}
\label{tab:llada_transfer}
\footnotesize
\setlength{\tabcolsep}{3.0pt}
\renewcommand{\arraystretch}{0.9}
\begin{tabular}{@{}lrrr@{}}
\toprule
Task & Vanilla Acc. & Fast valid & AUP \\
\midrule
GSM8K     & 79.53 & 79.98/4.21 & 332.40 \\
MATH-500  & 35.60 & 32.20/3.37 & 78.62 \\
HumanEval & 40.85 & 41.46/4.06 & 77.59 \\
MBPP-500  & 41.40 & 38.20/3.48 & 66.47 \\
\midrule
Average   & 49.35 & 47.96/3.78 & 138.77 \\
\bottomrule
\end{tabular}
\end{table}

The accelerated endpoints retain 47.96\% macro accuracy versus 49.35\% under
the conservative decode while reaching 3.78 average TPF; the full curves yield
138.77 macro AUP. This is substantially below the TAD-S-initialized primary
result (313.18 AUP and 7.87 TPF), consistent with the initializer supplying
much of the available cross-block parallelism. Because the recipes are not
matched, we do not assign this gap causally to initialization alone.

\paragraph{Operating-regime analysis.}
The transfer curve separates endpoint quality from available parallelism.
At confidence 0.9, its four-task macro operating point is 49.01\% accuracy at
2.71 TPF. Lowering the threshold to 0.8 increases average TPF by 26.3\% to
3.43, with a 1.05-point accuracy cost; selecting each task's fastest valid
endpoint further raises TPF by 10.4\% to 3.78 while leaving the macro accuracy
at 47.96\%. Thus, the learned policy supplies a useful threshold-controlled
speed range even when the student begins from vanilla LLaDA. The range is
nevertheless much narrower than for the TAD-S-initialized model. In particular,
the transfer endpoint is slightly more accurate than the primary operating
point (47.96\% versus 47.21\%) but reaches less than half its TPF (3.78 versus
7.87); endpoint accuracy alone therefore does not predict AUP.

The task curves also show where transfer is retained. GSM8K preserves nearly
80\% accuracy throughout the accepted range and contributes 332.40 AUP,
whereas MATH-500 and HumanEval contribute 78.62 and 77.59. MBPP reaches only
66.47 AUP, but this is closest to the primary model's 82.56 among the four
tasks in relative terms. Conversely, GSM8K has the largest absolute AUP gap
to the primary model (332.40 versus 848.31), despite its strong accuracy.
These differences suggest that the amount of compressible transition structure
available before OPD training is task dependent. Because the transfer recipe
changes several factors, this observation motivates a future
matched-initializer study rather than establishing a backbone effect.

\paragraph{Adaptive compression horizon.}
We finally isolate the scheduling benefit of adaptive compression from the
consistency verifier. Starting directly from LLaDA-8B-Instruct, without the
TAD-S initialization used elsewhere, we train matched fixed- and
adaptive-horizon policies under the same off-policy setup, disable verification
in both arms, and apply
the same certainty-forcing objective, which trains each committed token's confidence
toward the $0.9$ release threshold~\citep{chen2025dparallel}.
Table~\ref{tab:adaptive_horizon} reports four-benchmark macro results.

\begin{table}[t]
\centering
\caption{Single-seed fixed versus adaptive compression, with four-benchmark
macro Acc. and $\Delta$TPF relative to fixed.}
\label{tab:adaptive_horizon}
\small
\setlength{\tabcolsep}{3.0pt}
\renewcommand{\arraystretch}{0.85}
\begin{tabular}{rrrrrr}
\toprule
$H$ & \multicolumn{2}{c}{Acc.} & \multicolumn{2}{c}{TPF} & $\Delta$TPF \\
\cmidrule(lr){2-3}\cmidrule(lr){4-5}
 & Fixed & Adaptive & Fixed & Adaptive &  \\
\midrule
2 & 46.77 & \textbf{46.85} & 4.824 & \textbf{5.172} & +7.2\% \\
4 & \textbf{47.08} & 46.03 & 4.634 & \textbf{5.594} & +20.7\% \\
8 & \textbf{46.68} & 45.15 & 4.116 & \textbf{5.318} & +29.2\% \\
\bottomrule
\end{tabular}
\end{table}

Adaptive selection yields higher TPF at every matched horizon. At $H=2$,
macro accuracy is essentially unchanged (+0.08 points) while TPF increases by
7.2\%. Larger horizons trade 1.05--1.53 accuracy points for 20.7--29.2\%
more TPF. Fixed TPF instead falls from 4.824 to 4.116 as $H$ grows, so
adaptivity reaches operating points unattained by fixed horizons. It therefore
controls a state-dependent quality--efficiency trade-off rather than
unconditionally improving accuracy; the full method's consistency check is
the accuracy-oriented guardrail omitted here.

\section{Conclusion}
We presented \method{}, an on-policy transition-distillation framework for
few-step diffusion language models. It collects student-visited states and
uses a frozen, question-only teacher to construct state-dependent compressed
transitions without gold responses. Across reasoning and code-generation
tasks, it offers a favorable quality--efficiency trade-off. Matched
analyses support student-state supervision and adaptive transition horizons; a
decoder-controlled comparison shows 0.67-point higher macro accuracy and 4.4\%
higher average TPF beyond the common decoder. The guarantee covers teacher-verified
targets, not inference-time actions; confidence calibration and cross-backbone
transfer remain open. Improving both will be essential for translating
\method{}'s training-time gains into reliable acceleration across model
families, domains, and deployment settings. More broadly, \method{} suggests
that transition distillation should optimize not only token predictions, but
also the state-dependent scope of each parallel action. \method{} links learned
release scope to caching, verification, and decoding, enabling use in more
models, tasks, chips, systems, and domains. Joint systems evaluation is an
important next step. Future work should test whether these gains persist under
larger response budgets, diverse domains, and decoders with alternative cache
and release policies.

\newpage

\bibliography{references}

@inproceedings{austin2021d3pm,
  title={Structured Denoising Diffusion Models in Discrete State-Spaces},
  author={Austin, Jacob and Johnson, Daniel D. and Ho, Jonathan and Tarlow, Daniel and van den Berg, Rianne},
  booktitle={Advances in Neural Information Processing Systems},
  year={2021}
}

@inproceedings{hoogeboom2021argmax,
  title={Argmax Flows and Multinomial Diffusion: Learning Categorical Distributions},
  author={Hoogeboom, Emiel and Nielsen, Didrik and Jaini, Priyank and Forre, Patrick and Welling, Max},
  booktitle={Advances in Neural Information Processing Systems},
  year={2021}
}

@inproceedings{sahoo2024mdlm,
  title={Simple and Effective Masked Diffusion Language Models},
  author={Sahoo, Subham S. and Arriola, Marianne and Schiff, Yair and Gokaslan, Aaron and Marroquin, Edgar and Chiu, Justin T. and Rush, Alexander M. and Kuleshov, Volodymyr},
  booktitle={Advances in Neural Information Processing Systems},
  year={2024}
}

@inproceedings{arriola2025blockdiffusion,
  title={Block Diffusion: Interpolating Between Autoregressive and Diffusion Language Models},
  author={Arriola, Marianne and Gokaslan, Aaron and Chiu, Justin T. and Yang, Zhihan and Qi, Zhixuan and Han, Jiaqi and Sahoo, Subham Sekhar and Kuleshov, Volodymyr},
  booktitle={International Conference on Learning Representations},
  year={2025}
}

@article{nie2025llada,
  title={Large Language Diffusion Models},
  author={Nie, Shen and Zhu, Fengqi and You, Zebin and Zhang, Xiaolu and Ou, Jingyang and Hu, Jun and Zhou, Jun and Lin, Yankai and Wen, Ji-Rong and Li, Chongxuan},
  journal={arXiv preprint arXiv:2502.09992},
  year={2025}
}

@article{ye2025dream,
  title={Dream 7B: Diffusion Large Language Models},
  author={Ye, Jiacheng and Xie, Zhihui and Zheng, Lin and Gao, Jiahui and Wu, Zirui and Jiang, Xin and Li, Zhenguo and Kong, Lingpeng},
  journal={arXiv preprint arXiv:2508.15487},
  year={2025}
}

@article{cheng2025sdar,
  title={SDAR: A Synergistic Diffusion-Autoregression Paradigm for Scalable Sequence Generation},
  author={Cheng, Shuang and Bian, Yihan and Liu, Dawei and Jiang, Yuhua and Liu, Yihao and Zhang, Linfeng and Wang, Wenhai and Guo, Qipeng and Chen, Kai and Qi, Biqing and Zhou, Bowen},
  journal={arXiv preprint arXiv:2510.06303},
  year={2025}
}

@article{wu2025fastdllm,
  title={Fast-dLLM: Training-free Acceleration of Diffusion LLM by Enabling KV Cache and Parallel Decoding},
  author={Wu, Chengyue and Zhang, Hao and Xue, Shuchen and Liu, Zhijian and Diao, Shizhe and Zhu, Ligeng and Luo, Ping and Han, Song and Xie, Enze},
  journal={arXiv preprint arXiv:2505.22618},
  year={2025}
}

@article{chen2025dparallel,
  title={dParallel: Learnable Parallel Decoding for dLLMs},
  author={Chen, Zigeng and Fang, Gongfan and Ma, Xinyin and Yu, Ruonan and Wang, Xinchao},
  journal={arXiv preprint arXiv:2509.26488},
  year={2025}
}

@inproceedings{wang2026d2f,
  title={Diffusion LLMs Can Do Faster-Than-AR Inference via Discrete Diffusion Forcing},
  author={Wang, Xu and Xu, Chenkai and Jin, Yijie and Jin, Jiachun and Zhang, Hao and Deng, Zhijie},
  booktitle={International Conference on Learning Representations},
  year={2026}
}

@article{d3llm2026,
  title={d3LLM: Ultra-Fast Diffusion LLM using Pseudo-Trajectory Distillation},
  author={Qian, Yu-Yang and Su, Junda and Hu, Lanxiang and Zhang, Peiyuan and Deng, Zhijie and Zhao, Peng and Zhang, Hao},
  journal={arXiv preprint arXiv:2601.07568},
  year={2026}
}

@inproceedings{cdlm2026,
  title={CDLM: Consistency Diffusion Language Models for Faster Sampling},
  author={Kim, Minseo and Xu, Chenfeng and Hooper, Coleman and Singh, Harman and Athiwaratkun, Ben and Zhang, Ce and Keutzer, Kurt and Gholami, Amir},
  booktitle={Proceedings of the 9th Conference on Machine Learning and Systems},
  year={2026}
}

@article{cd4lm2026,
  title={CD4LM: Consistency Distillation and Adaptive Decoding for Diffusion Language Models},
  author={Liang, Yihao and Wang, Ze and Chen, Hao and Sun, Ximeng and Wu, Jialian and Yu, Xiaodong and Liu, Jiang and Barsoum, Emad and Liu, Zicheng and Jha, Niraj K.},
  journal={arXiv preprint arXiv:2601.02236},
  year={2026}
}

@inproceedings{speed2026,
  title={SPEED: Sharpened-Teacher Distillation for Parallel Decoding of Diffusion Language Models},
  author={Shen, Qiuhong and Yang, Xingyi and Ma, Xinyin and Fang, Gongfan and Wang, Xinchao},
  booktitle={Proceedings of the 43rd International Conference on Machine Learning},
  year={2026}
}

@article{tad2026,
  title={TAD: Temporal-Aware Trajectory Self-Distillation for Fast and Accurate Diffusion LLM},
  author={Zhou, Haoyang and Kong, Li and Ren, Shijie and Wang, Xiting and Liang, Shuang and Wang, Guowei and Pan, Zhenxuan},
  journal={arXiv preprint arXiv:2605.09536},
  year={2026}
}

@article{t3d2026,
  title={T3D: Few-Step Diffusion Language Models via Trajectory Self-Distillation with Direct Discriminative Optimization},
  author={Zhang, Tunyu and Zhang, Xinxi and Han, Ligong and Shi, Haizhou and He, Xiaoxiao and Li, Zhuowei and Wang, Hao and Xu, Kai and Srivastava, Akash and Pavlovic, Vladimir and Metaxas, Dimitris N.},
  journal={arXiv preprint arXiv:2602.12262},
  year={2026}
}

@article{luo2026dopsd,
  title={Learning from the Self-Future: On-Policy Self-Distillation for dLLMs},
  author={Luo, Yifu and Chen, Zeyu and Wang, Haoyu and Hu, Xinhao and Zhang, Yuxuan and Sha, Zhizhou and Liu, Shiwei},
  journal={arXiv preprint arXiv:2606.18195},
  year={2026}
}

@article{su2026opdlm,
  title={Data-Efficient Autoregressive-to-Diffusion Language Models via On-Policy Distillation},
  author={Su, Xingyu and Helwig, Jacob and Parashar, Shubham and Chagi, Atharv and Jotsna, Lakshmi and Zhi, Degui and Caverlee, James and Kalathil, Dileep and Ji, Shuiwang},
  journal={arXiv preprint arXiv:2606.06712},
  year={2026}
}

@article{agarwal2023gkd,
  title={On-Policy Distillation of Language Models: Learning from Self-Generated Mistakes},
  author={Agarwal, Rishabh and Vieillard, Nino and Zhou, Yongchao and Stanczyk, Piotr and Ramos, Sabela and Geist, Matthieu and Bachem, Olivier},
  journal={arXiv preprint arXiv:2306.13649},
  year={2023}
}

@inproceedings{salimans2022progressive,
  title={Progressive Distillation for Fast Sampling of Diffusion Models},
  author={Salimans, Tim and Ho, Jonathan},
  booktitle={International Conference on Learning Representations},
  year={2022}
}

@inproceedings{song2023consistency,
  title={Consistency Models},
  author={Song, Yang and Dhariwal, Prafulla and Chen, Mark and Sutskever, Ilya},
  booktitle={International Conference on Machine Learning},
  year={2023}
}

@article{cobbe2021gsm8k,
  title={Training Verifiers to Solve Math Word Problems},
  author={Cobbe, Karl and Kosaraju, Vineet and Bavarian, Mohammad and Chen, Mark and Jun, Heewoo and Kaiser, Lukasz and Plappert, Matthias and Tworek, Jerry and Hilton, Jacob and Nakano, Reiichiro and Hesse, Christopher and Schulman, John},
  journal={arXiv preprint arXiv:2110.14168},
  year={2021}
}

@article{hendrycks2021math,
  title={Measuring Mathematical Problem Solving with the MATH Dataset},
  author={Hendrycks, Dan and Burns, Collin and Kadavath, Saurav and Arora, Akul and Basart, Steven and Tang, Eric and Song, Dawn and Steinhardt, Jacob},
  journal={arXiv preprint arXiv:2103.03874},
  year={2021}
}

@article{chen2021humaneval,
  title={Evaluating Large Language Models Trained on Code},
  author={Chen, Mark and Tworek, Jerry and Jun, Heewoo and Yuan, Qiming and Pinto, Henrique Ponde de Oliveira and Kaplan, Jared and Edwards, Harri and Burda, Yuri and Joseph, Nicholas and Brockman, Greg and Ray, Alex and Puri, Raul and Krueger, Gretchen and Petrov, Michael and Khlaaf, Heidy and Sastry, Girish and Mishkin, Pamela and Chan, Brooke and Gray, Scott and Ryder, Nick and Pavlov, Mikhail and Power, Alethea and Kaiser, Lukasz and Bavarian, Mohammad and Winter, Clemens and Tillet, Philippe and Such, Felipe Petroski and Cummings, Dave and Plappert, Matthias and Paino, Alex and Tezak, Nikolas and Tang, Jie and Babuschkin, Igor and Balaji, Suchir and Jain, Shantanu and Saunders, William and Hesse, Christopher and Carr, Andrew N. and Leike, Jan and Achiam, Josh and Misra, Vedant and Morikawa, Evan and Knight, Matthew and Brundage, Miles and Murati, Mira and Mayer, Katie and Welinder, Peter and McGrew, Bob and Amodei, Dario and McCandlish, Sam and Sutskever, Ilya and Zaremba, Wojciech},
  journal={arXiv preprint arXiv:2107.03374},
  year={2021}
}

@article{austin2021programsynthesis,
  title={Program Synthesis with Large Language Models},
  author={Austin, Jacob and Odena, Augustus and Nye, Maxwell and Bosma, Maarten and Michalewski, Henryk and Dohan, David and Jiang, Ellen and Cai, Carrie and Terry, Michael and Le, Quoc and Sutton, Charles},
  journal={arXiv preprint arXiv:2108.07732},
  year={2021}
}

@inproceedings{hu2021lora,
  title={LoRA: Low-Rank Adaptation of Large Language Models},
  author={Hu, Edward J. and Shen, Yelong and Wallis, Phillip and Allen-Zhu, Zeyuan and Li, Yuanzhi and Wang, Shean and Wang, Lu and Chen, Weizhu},
  booktitle={International Conference on Learning Representations},
  year={2022}
}

@article{ren2026topd,
  title={Trace-Based On-Policy Distillation for Masked Diffusion Language Models},
  author={Ren, Haolin and Huang, Ziyang and Yuan, Chenhao and Zhao, Jun and Liu, Kang},
  journal={arXiv preprint arXiv:2607.16872},
  year={2026}
}

\end{document}